\documentclass[11pt]{article}

\usepackage[final]{acl}

\usepackage{times}
\usepackage{latexsym}
\usepackage{booktabs}
\usepackage[table]{xcolor} 
\usepackage{arydshln}
\usepackage{makecell}
\usepackage{multirow}
\usepackage{amsmath}
\usepackage{mathtools}
\usepackage{amssymb}
\usepackage{mathtools}
\usepackage{amsthm}
\usepackage{tabularx}
\newcolumntype{Y}{>{\centering\arraybackslash}X} 
\usepackage{booktabs}
\usepackage{url}        
\usepackage{algpseudocode}
\usepackage{algorithm}
\usepackage{algorithmicx}

\usepackage{amsthm}
\DeclareMathOperator*{\argmin}{argmin}

\definecolor{softyellow}{RGB}{253,231,111} %
\definecolor{softred}{RGB}{231,76,60} 
\definecolor{fastgreen}{RGB}{80,200,120} 

\usepackage[T1]{fontenc}

\usepackage[utf8]{inputenc}

\usepackage{microtype}

\usepackage{inconsolata}

\usepackage{graphicx}

\title{FastKV: Decoupling of Context Reduction and KV Cache Compression\\for Prefill-Decoding Acceleration}

\author{
 \textbf{Dongwon Jo\textsuperscript{1}\thanks{Equal Contribution}},
 \textbf{Jiwon Song\textsuperscript{1}\footnotemark[1]},
 \textbf{Yulhwa Kim\textsuperscript{2}},
 \textbf{Jae-Joon Kim\textsuperscript{1}}
\\
 \textsuperscript{1}Seoul National University,
 \textsuperscript{2}Sungkyunkwan University
\\
\texttt{\{dongwonjo,jiwon.song,kimjaejoon\}@snu.ac.kr} \\
\texttt{yulhwakim@skku.edu}\\
}

\begin{document}
\maketitle

\begin{abstract}
While large language models (LLMs) excel at handling long-context sequences, they require substantial prefill computation and key-value (KV) cache, which can heavily burden computational efficiency and memory usage in both prefill and decoding stages.
Recent works that compress KV caches with prefill acceleration reduce this cost but inadvertently tie the prefill compute reduction to the decoding KV budget. This coupling arises from overlooking the layer-dependent variation of critical context, often leading to accuracy degradation. 
To address this issue, we introduce FastKV, a KV cache compression framework designed to reduce latency in both prefill and decoding by leveraging the stabilization of token importance in later layers.
FastKV performs full-context computation until a Token-Selective Propagation (TSP) layer, which forwards only the most informative tokens to subsequent layers.
From these propagated tokens, FastKV independently selects salient KV entries for caching, thereby decoupling KV budget from the prefill compute reduction based on the TSP decision.
This independent control of the TSP rate and KV retention rate enables flexible optimization of efficiency and accuracy.
Experimental results show that FastKV achieves speedups of up to 1.82$\times$ in prefill and 2.87$\times$ in decoding compared to the full-context baseline, while matching the accuracy of the decoding-only baselines.
Our code is available at \url{https://github.com/dongwonjo/FastKV}.

\end{abstract}

\section{Introduction}

Large Language Models (LLMs) have rapidly advanced and now support extended context windows of 128K and even beyond one million tokens~\citep{gpt4,gemini,claude3}.
This capability significantly enables a broad range of applications for LLMs such as retrieval-augmented generation, multi-document reasoning, and code generation~\citep{intro1,intro2,intro3}. 
However, the computational and memory overhead of long-context inference remains a critical bottleneck. 

Long-context inference introduces substantial burdens in both the prefill and decoding stages.
In the prefill stage, attention computation scales quadratically with input length, making very long prompts expensive to process.
In the decoding stage, the large amount of KV cache becomes the dominant factor, consuming GPU memory and reducing throughput since every generated token must repeatedly access this cache.
Together, these trends make inference prohibitively costly: prefill slows down with longer inputs, while decoding efficiency deteriorates as the cache size grows with context length.

\begin{table}[t]
\centering
\scalebox{0.76}{
\renewcommand{\arraystretch}{0.9}
\setlength{\tabcolsep}{14pt} 
\begin{tabular}{lccc}
\toprule
\textbf{Method} & \textbf{Prefill} & \textbf{Decoding} & \textbf{Acc.} \\
\midrule
Full-context    & \cellcolor{softred!70}{Slow}   & \cellcolor{softred!70}{Slow}     & \cellcolor{fastgreen!70}{High} \\
\midrule
StreamingLLM     & \cellcolor{softred!70}{Slow}   & \cellcolor{fastgreen!70}{Fast}  & \cellcolor{softred!70}{Low} \\
SnapKV     & \cellcolor{softred!70}{Slow}   & \cellcolor{fastgreen!70}{Fast}  & \cellcolor{fastgreen!70}{High} \\
\midrule
GemFilter  & \cellcolor{fastgreen!70}{Fast}  & \cellcolor{fastgreen!70}{Fast}  & \cellcolor{softred!70}{Low} \\
\textbf{FastKV}     & \cellcolor{fastgreen!70}{\textbf{Fast}}  & \cellcolor{fastgreen!70}{\textbf{Fast}}  & \cellcolor{fastgreen!70}{\textbf{High}} \\
\bottomrule
\end{tabular}
} 
\caption{Comparison of KV cache compression methods. FastKV uniquely achieves fast prefill and decoding, and high accuracy simultaneously.}
\vspace{-5mm}
\label{tab1:comparison}
\end{table}

Recent studies have explored two complementary directions to overcome these burdens. 
Most existing KV cache compression methods, such as StreamingLLM~\cite{streamingllm} and SnapKV~\citep{snapkv} target the decoding stage, by pruning already-generated KV cache, but do not accelerate the prefill stage at all.

In contrast, GemFilter~\citep{gemfilter} and PyramidInfer~\citep{pyramidinfer} focus on the prefill stage, aiming to reduce the quadratic cost of processing long prompts by generating KV cache of only critical tokens. 
Despite these advances, a fundamental trade-off remains: decoding-focused approaches fail to alleviate the prefill burden, whereas prefill-focused methods compromise accuracy when the KV budget is reduced to levels that would significantly accelerate decoding.

To bridge this gap, we propose FastKV, a KV cache compression framework that accelerates prefill and decoding stages without compromising accuracy.
FastKV is motivated by two key observations: (i) during prefill, early layers must propagate the full-context so that later layers retain the opportunity to attend to any part of the context; (ii) however, during decoding, each layer ultimately attends to only a small fraction of the prefilled tokens, meaning that it is unnecessary to retain the entire KV cache built during prefill.

Building on these insights, FastKV introduces two techniques.
First, we adopt a two-stage prefill strategy that retains the full-context in early layers while context usage remains unstable, and, once stabilization becomes evident, switches to propagating only salient tokens in later layers.
Second, we decouple the KV budget to separate the context processed during prefill from the amount of KV cache retained for decoding.

As summarized in Table~\ref{tab1:comparison}, these techniques address the prefill–decoding trade-off by performing context reduction at the right time and preserving only critical KV caches for decoding, achieving up to 1.82$\times$ faster prefill and up to 2.87$\times$ faster decoding compared to the full-context baseline, while maintaining accuracy drop within 1\% on LongBench benchmark.

\section{Background}

\subsection{Bottlenecks of Long-context Inference}

Auto-regressive LLM inference consists of two stages: prefill and decoding stage.

\begin{itemize}
\item In the prefill stage, the model processes the entire input prompt and builds the KV cache across all layers. 
The computational cost of attention in this stage scales quadratically with the input length.

\item In the decoding stage, the model generates tokens auto-regressively while reusing the KV cache. 
Here, the per-step attention cost is only linear in the number of cached tokens, but the cache itself grows linearly with input length and must be repeatedly accessed at every step.
\end{itemize}

\noindent When the context is short, these costs are manageable. 
However, as context length increases, both stages become severe bottlenecks: 
Prefill latency explodes as the context length increases due to quadratic attention, and the decoding latency deteriorates as the linearly growing KV cache must be repeatedly accessed at every step, creating significant memory bandwidth overhead.

\subsection{Prior Approaches and Limitations}

A large body of work has explored KV cache compression to alleviate the burden of long-context inference.
Early efforts such as StreamingLLM~\citep{streamingllm} exploit the observation of attention sink tokens, retaining only those sink tokens together with the most recent context in the KV cache.
Building on this idea, SnapKV and H2O~\citep{snapkv,h2o} introduced attention-based importance metrics to retain only salient tokens, thereby reducing the KV cache more selectively.
Subsequent studies~\citep{adakv,headkv,pyramidkv} further refined this direction by assigning fine-grained KV budgets at the head or layer level, aiming to preserve accuracy.
However, these methods provide at best marginal accuracy improvements over SnapKV, while leaving the fundamental bottleneck unsolved: they still require producing the KV cache for the  full-context before selecting which tokens to retain, so prefill latency remains unreduced.

In contrast, prefill-aware KV cache compression methods have emerged.
Instead of processing all tokens during prefill, they attempt to accelerate inference by reducing the effective context length up front.
For example, GemFilter~\citep{gemfilter} leverages pre-defined filter layer's attention to select a compact subset of input tokens, while PyramidInfer~\citep{pyramidinfer} exploits cross-layer redundancy to gradually reduce the hidden states propagated to subsequent layers.
These approaches naturally produce a smaller KV cache, yielding both prefill and decoding speedups.
Nevertheless, they suffer from inherent trade-offs: GemFilter enforces the same reduced set of tokens across all subsequent layers, discarding potentially useful information for deeper processing, while PyramidInfer, though layer-aware, still prunes aggressively from early layers, both of which compromise accuracy.
Moreover, in these designs, the KV cache compression is tightly coupled with the amount of prefill compute reduction: achieving sufficient decoding acceleration requires aggressive context pruning, which simultaneously amplifies accuracy degradation.
Importantly, this fragility is not due to a tighter memory budget; prefill-aware schemes drop tokens on the fly, blocking later layers from attending to them and thus degrading accuracy.
The drawbacks of existing prefill-aware KV cache compressions are further discussed in Appendix~\ref{appendix:distinction}.


These limitations suggest that existing methods are inherently constrained by conflating KV cache compression with prefill compute reduction into a single stage, and a more balanced approach is needed to jointly optimize prefill and decoding efficiency.

\section{Motivations}

\subsection{Layer-dependent Context Dynamics}
\label{subsection:context_dynamics}

A key challenge in reducing prefill latency is to determine when to shrink the amount of context processed at each layer.
Existing approaches present two contrasting strategies: GemFilter processes the context up to the filter layer to select salient tokens and restart the prefill stage with the selected tokens, while PyramidInfer gradually reduces the context size layer by layer.

In both methods, even early layers lose the opportunity to access full-context of the tokens.
They do not directly examine how the selection of critical tokens evolves across layers, and to what extent early pruning may disrupt later layers’ ability to attend to their eventual targets. 
To answer this question, we analyze how the layer-wise critical tokens changes as depth increases.

\begin{figure}[t]
    \centering
    \includegraphics[width=1.0\columnwidth]{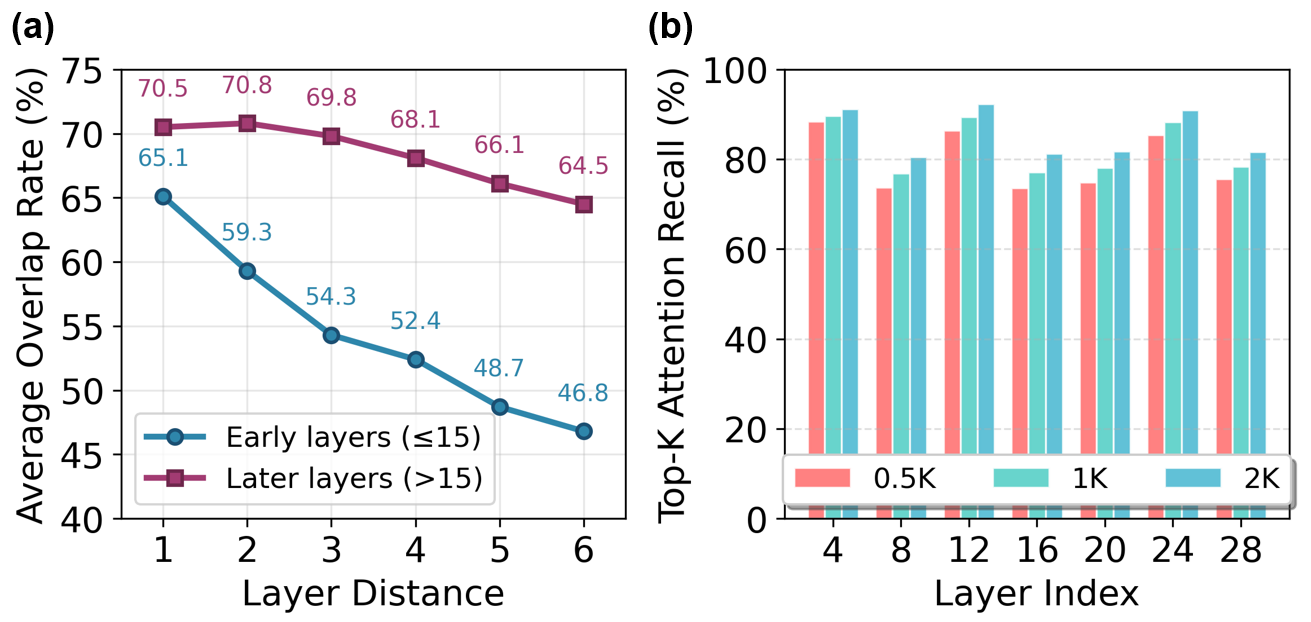}
    \vspace{-7mm}
    \caption{(a) Early layers exhibit unstable context focus, reflected by low critical token overlap. (b) Attention distributions are sparse, with Top-K tokens dominating the scores.}
    \label{fig1}
    \vspace{-4mm}
\end{figure}

We feed a 128K-token input into LLaMA-3.1-8B-Instruct and, at each layer, collect the top-512 critical tokens that receive the highest average attention mass across heads. 
We then calculate the average overlap ratio of these critical token indices between layers as the layer distance increases. 
Figure~\ref{fig1}(a) presents the results. 
In the early layers ($\leq$15), the overlap ratio drops sharply with increasing layer distance, indicating that the critical tokens perceived by each layer shift sharply. 
In contrast, in the later layers ($>$15), the overlap decays much more slowly, suggesting that the same subset of tokens remains consistently important across multiple successive layers.

These observations highlight the difference in context utilization across layers. 
In the early layers, attention focus is highly unstable, and pruning tokens at this stage irreversibly remove tokens that later layers would otherwise consider critical. 
Once discarded, these tokens cannot be recovered, causing downstream layers to lose access to potentially indispensable context and leading to severe performance degradation. 
In contrast, in the later layers, the set of critical tokens shows a high degree of overlap across layers, so aggressive token pruning can be applied with minimal impact on the model accuracy.
This layer-dependent context dynamic implies that token pruning during the prefill stage must process the full-context in early layers, and then transition to selective context propagation in later layers.

\begin{figure*}[t]
    \centering
    \includegraphics[width=1.0\textwidth]{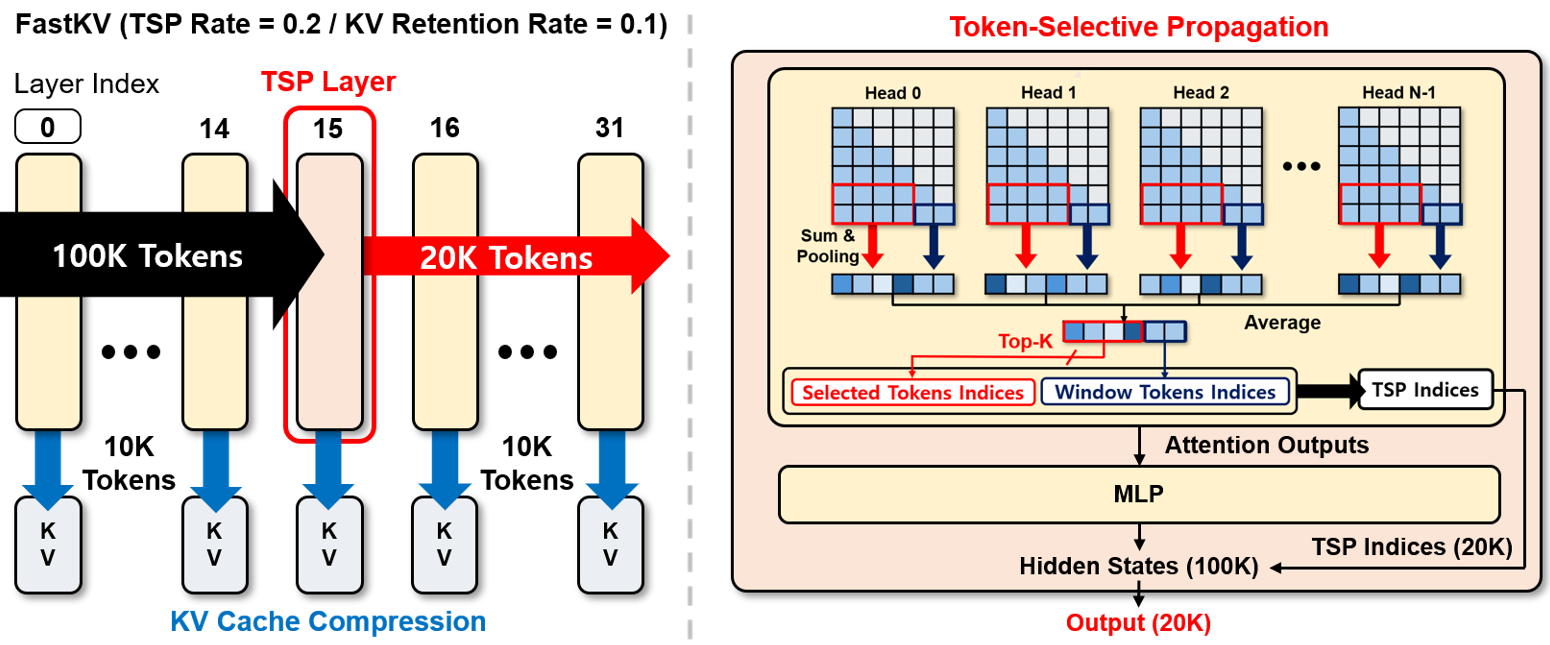}
    \vspace{-6mm}
    \caption{Illustration of the proposed FastKV scheme. The proposed FastKV introduce Token-Selective Propagation approach to selectively propagate only a limited set of tokens while effectively compressing KV cache.}
    \label{fig2}
    \vspace{-4mm}
\end{figure*}

\subsection{Sparse Context Utilization Across Layers}
The results of Section~\ref{subsection:context_dynamics} suggest that early layers must be allowed to process the full-context during prefill so that later layers do not lose access to potentially critical tokens. 
However, this does not imply that each layer has to cache all of the KV values that it generated.
To investigate what fraction of the context is actually used during decoding, we measure the top-$K$ attention recall: the proportion of total attention mass covered by the $K$ most attended tokens at each layer.

As shown in Figure~\ref{fig1}(b), across all layers of LLaMA-3.1-8B-Instruct with 128K input tokens, only a sparse subset of tokens dominates the attention distribution. 
Even with K = 512 (0.38\% of entire tokens), the majority of the attention mass is already captured.
This indicates that all layers only rely on a small subset of tokens once decoding begins.

This phenomenon of sparse utilization is consistent with results from prior works on KV cache compression during decoding stage, such as SnapKV and H2O. These studies have demonstrated that KV cache compression can be highly effective because only a small fraction of tokens are actively used during decoding. 

Overall, our analysis suggests that the early layers require careful management of KV values. In the prefill stage, the full-context must be processed to provide subsequent layers with complete contextual information. However, the model can cache only a small subset of generated tokens, as the decoding stage depends on only a small fraction of the prefill context.

\section{Proposed FastKV}

\subsection{Overview of FastKV}

The overall workflow of FastKV is illustrated in Figure~\ref{fig2}. 
FastKV accelerates long-context inference by rethinking how much context is propagated during prefill and how much KV cache is retained during decoding.
FastKV introduces two complementary innovations:

\begin{itemize}
\item \textbf{Token-Selective Propagation (TSP)}. A dedicated TSP layer, placed around the middle of the decoder, forwards only a selected subset of hidden states to later layers rather than propagating the entire prompt. 
This reduces the context passed downstream while keeping critical information intact.

\item \textbf{Layer-wise KV retention}. During prefill, each layer independently discards less influential entries and preserves only a specified retention rate of its KV cache. 
After prefill, every layer thus maintains a compressed KV cache, which significantly accelerates decoding without degrading accuracy.
\end{itemize}

\noindent By allowing early layers to process the full-context before compression, FastKV ensures that they can freely identify the tokens to retain, while later layers, which tend to converge on similar subsets, remain robust even when operating on reduced context. This design allows every layer to carry a compressed but meaningful KV cache into decoding.

Compared to prior work, FastKV avoids the rigid constraints of GemFilter and PyramidInfer. 
GemFilter enforces a single layer’s token selection across all layers, which is particularly harmful to early layers where each layer attends to different subsets of tokens.
PyramidInfer reduces context from the very first layers, limiting the flexibility of each layer to select its own important tokens. Furthermore, both methods couple prefill compression with the decoding KV budget, whereas FastKV decouples them, enabling independent control of how much context is propagated during prefill and how much KV is preserved for decoding. This flexibility yields a superior accuracy–efficiency trade-off.

\subsection{Two-stage Prefill with Token-Selective Propagation}

As shown in Section~\ref{subsection:context_dynamics}, the set of critical tokens fluctuates substantially in the early layers but stabilizes in later layers.
This observation motivates a two-stage prefill strategy: using full-context computation in the early layers to capture diverse token dependencies, and then reducing the context in subsequent layers once token importance has stabilized.

To implement this, FastKV introduces a dedicated TSP layer. 
In this layer, each token is evaluated by how strongly it is attended by the recent window tokens, which serve as queries.  
Specifically, for each token $i$, its saliency score $S$ is computed by averaging attention weights over all heads when queried from the window tokens:

\begin{equation}
    \label{eq:TSP-1}
    S^{l,h}_{i}=\text{Pooling}(\sum_{n=0}^{N_{\text{obs}}} Att_l[h,\,N_I-n,\,i+m])
\end{equation} 

\begin{equation}
    \label{eq:TSP-2}
    S_{i}^{TSPlayer}=\frac{1}{H}\sum_{h=0}^{H-1}S_i^{TSPlayer,h},
\end{equation} 

Here, $S_i^{l,h}$ is the saliency score of $i$-th token in $h$-th attention head of the $l$-th layer. $Att_l$ denotes the attention score matrix of $l$-th layer, while $N_I$ and $N_{obs}$ indicate the number of tokens in the input prompt and the window token size, respectively. $H$ denotes the number of attention heads in each layer. Since compressing the layer output requires a single index set, we average the attention weights to calculate a score $S_{i}^{TSPlayer}$ that represents the saliency of tokens at the layer level, as outlined in Equation~\ref{eq:TSP-2}.


Based on these saliency scores, indices of the top-ranked tokens up to the predefined TSP rate are selected.  
Crucially, all window tokens themselves are always included in the propagated set, and their indices are merged with the saliency-selected ones to form the final TSP token set passed to the next layer. 

Prefill therefore proceeds in two distinct stages. 
In the first stage, from the input up to the TSP layer, all layers process the full-context and construct their compressed KV caches.
In the second stage, starting at the TSP layer, only the reduced subset of saliency-selected tokens (together with all window tokens) continues forward, and later layers process this compressed context to form their compressed KV caches.

By applying Token-Selective Propagation only after stabilization, FastKV preserves the heterogeneous attention patterns of early layers while still reducing prefill latency in later layers.

\subsection{Impact of TSP and TSP Layer Selection}

\begin{figure}[t]
    \centering
    \includegraphics[width=0.90\columnwidth]{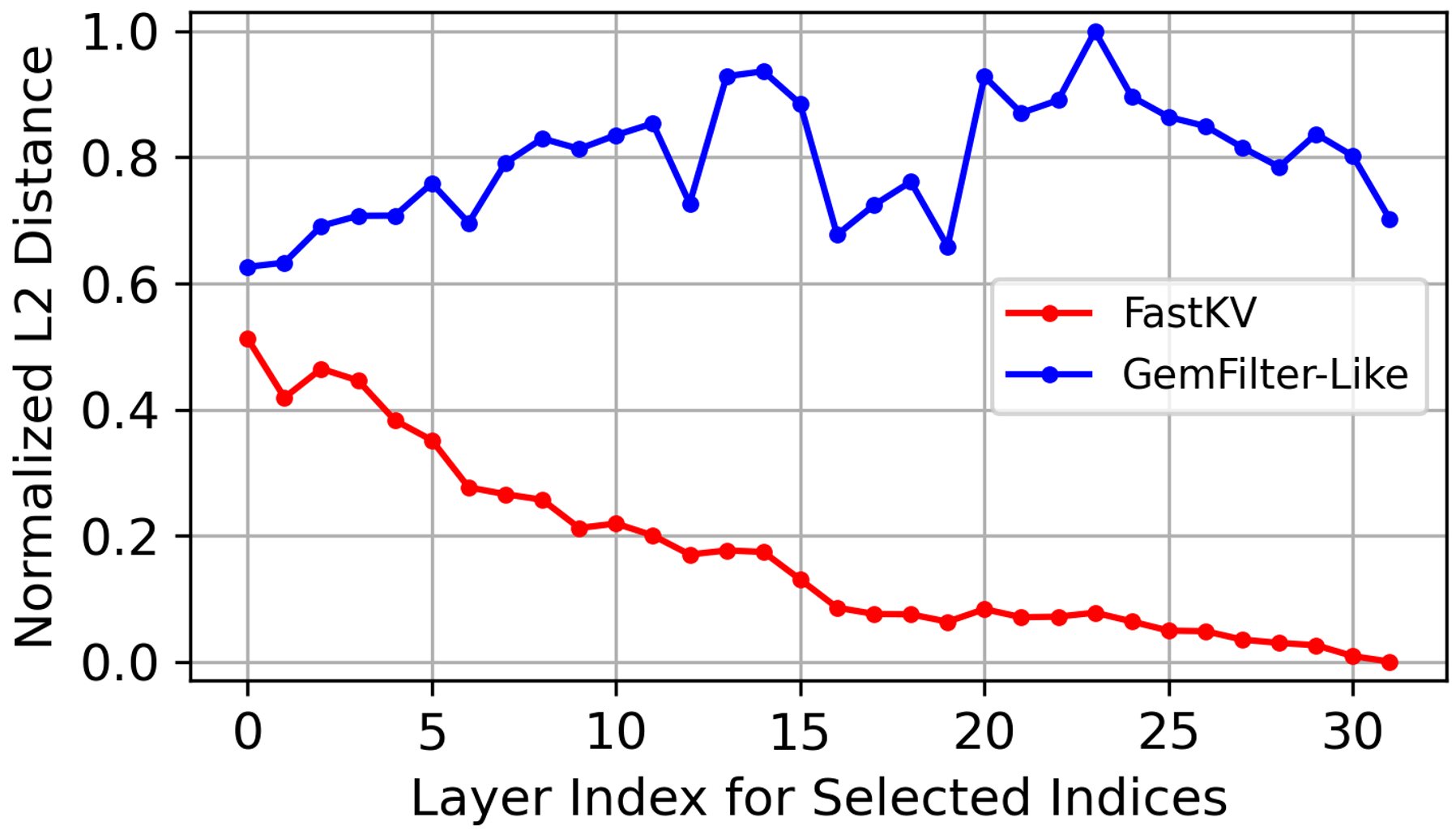}
    \caption{Comparison of normalized L2 distances between hidden states generated by the full-context baseline, TSP, and GemFilter-like methods.}
    \label{fig3}
    \vspace{-4mm}
\end{figure}

We first investigate the effect of applying Token-Selective Propagation (TSP) at different layers. 
For each candidate TSP layer, we compute the final logits and compare them against those obtained from the full-context baseline. 
Figure~\ref{fig3} reports the normalized L2 distance between the two outputs for LLaMA-3.1-8B-Instruct.

We also include a GemFilter-like strategy as a baseline, where a single filter layer selects a subset of tokens and then the model is re-prefilled using only this subset. 
This procedure forces even the early layers, which normally attend to distinct tokens, to process the same limited token set, resulting in significant deviations in the final logits.

In contrast, TSP preserves full-context processing before the TSP layer, so each early layer can still attend to its own preferred subset of tokens. 
As a result, when the TSP layer is placed at the middle or later part of the model, the logits remain much closer to the full-context baseline than those produced by GemFilter, as shown in Figure~\ref{fig3}.

If the TSP layer is placed too early, token selection becomes overly restricted and the resulting logits deviate substantially from the full-context baseline. 
If it is placed too late, many layers still process the full-context and prefill latency reduction is limited. 
Therefore, selecting the TSP layer at the right position is critical, which we formalize as follows:

\vspace{-5mm}
\begin{equation} 
\label{eq:TSP-selection} 
L_{TSP} = \argmin_{L \leq L_{\max}} \frac{1}{N} \sum_{i=1}^{N} \left\| \mathbf{H}_i - \mathbf{H}'_{L,i} \right\|_2^2,
\end{equation}

where $\mathbf{H}_i$ denotes the hidden state at the final layer under full-context for the $i$-th calibration input, and $\mathbf{H}'_{L,i}$ the corresponding hidden state when TSP is applied at candidate layer $L$.

The criterion searches for the earliest layer whose output remains close to the full-context baseline, while constraining $L \leq L_{\max}$ to avoid excessively late placement. 
This ensures that the chosen TSP layer simultaneously minimizes model degradation and provides tangible prefill latency savings.

\subsection{KV Retention Decoupled with Prefill Compute}

In prefill-aware methods such as GemFilter and PyramidInfer, the amount of context reduced during prefill directly determines the KV cache size available for decoding.
This tight coupling means that reducing the KV budget inevitably requires more aggressive prefill compute reduction. 
As a result, early layers are deprived of tokens they uniquely rely on, leading to significant accuracy degradation.

FastKV breaks this coupling by introducing two independent hyperparameters: the TSP rate, which controls how much context is propagated after the TSP layer during prefill, and the KV retention rate, which specifies how many of each layer’s KV entries are preserved for reuse during decoding. 


By decoupling prefill compute reduction from decoding KV retention, FastKV allows latency savings in prefill and memory savings in decoding to be tuned separately. For example, one can adopt a conservative TSP rate to protect accuracy in prefill, while still using a tighter KV retention rate to reduce KV size during decoding. 
Conversely, when memory constraints are critical, both rates can be reduced together to maximize efficiency. 
This independent control enables FastKV to achieve better accuracy–efficiency trade-offs than GemFilter and PyramidInfer, which couple the two stages and thus sacrifice accuracy when aiming for stronger KV cache compression.

\section{Experiments}

\subsection{Setup}
\label{subsec:setup}
\textbf{Models and Datasets.} 
We evaluate two open-source LLMs of different sizes: LLaMA-3.1-8B-Instruct~\citep{llama3} and Ministral-8B-Instruct~\citep{ministral}.
These models have 32 and 36 decoder layers, respectively, and employ GQA~\citep{gqa} with a context window size of 128K tokens.
More evaluation results on an additional model are provided in Appendix~\ref{appendix:more_experimental_results}
We use LongBench~\citep{longbench}, which consists of 16 subtasks across single-document QA, multi-document QA, summarization, few-shot learning, synthetic tasks, and code completion to assess the models' long-context understanding capabilities.
For further examination, LLaMA-3.1-8B-Instruct is also evaluated on Needle-in-a-Haystack~\citep{needle} and RULER~\citep{ruler}.
See Appendix~\ref{appendix:evaluation_details} for more details on evaluation.

\noindent\textbf{Implementation Details.}
We integrate our proposed FastKV method upon self-attention implementation of HuggingFace Transformers library, which utilizes FlashAttention-2~\citep{flashattention2} kernel. We select layer 15 as the TSP layer for LLaMA-3.1-8B-Instruct and layer 17 for Ministral-8B-Instruct. We fix the observation window size to 8 and the pooling kernel size to 7. The TSP rate is set to 20\% for evaluation. The setting yields 60\% prefill compute rate for both models.

\noindent\textbf{Baselines.}
We compare FastKV against five baseline methods for KV cache compression: StreamingLLM~\citep{streamingllm}, H2O~\citep{h2o} and SnapKV~\citep{snapkv}(decoding-only acceleration); PyramidInfer~\citep{pyramidinfer} and GemFilter~\citep{gemfilter} (prefill-aware acceleration).
For all methods except PyramidInfer, results are reported at both 10\% and 20\% KV retention rates.
The indices of GemFilter filter layers for LLaMA-3.1-8B-Instruct and Ministral-8B-Instruct are 13 and 17, respectively.
For GemFilter, this yields 51\%  and 61\% of prefill compute rate for LLaMA.
This setting yields 60\% and 70\% of prefill compute rate for Ministral.
Setting prefill compute rate 60\% for PyramidInfer automatically determines its KV retention rate to be 60\%, which is equal to its prefill compute rate.

\subsection{Accuracy}
\label{subsec:accuracy}

\textbf{LongBench.}
The accuracy evaluation results on LongBench are summarized in Table~\ref{tab:longbench}. The full breakdown of results is provided in Appendix~\ref{appendix:more_experimental_results_longbench}.
Previous works for decoding-only acceleration, such as StreamingLLM and H2O, show significant accuracy degradation, whereas SnapKV maintains accuracy after KV cache compression,  with average drops below 1.46\% and 0.76\% at 10\% and 20\% KV retention, respectively.
GemFilter recomputes the prefill stage on a fragmented input consisting only of selected tokens, thereby discarding information carried by the removed tokens and incurring substantially larger accuracy drop of up to 11.58\%.
PyramidInfer, even under a higher KV retention rate, begins prefill compute reduction from early layers where context usage is still unstable and stores the resulting compressed KV, which propagates information loss and yields inferior accuracy.
By contrast, FastKV attains memory and computational efficiency comparable to GemFilter while preserving accuracy on par with the full-context baseline.
This demonstrates the effectiveness of FastKV’s two-stage prefill strategy retaining the full-context in early layers and propagating only salient tokens in later layers.

\begin{table*}[ht]
\centering
\caption{LongBench results on LLaMA-3.1-8B-Instruct and Ministral-8B-Instruct.}
\label{tab:longbench}
\resizebox{0.9\textwidth}{!}{
\renewcommand{\arraystretch}{0.82}
\setlength{\tabcolsep}{1.4pt}
\scalebox{0.70}{
\begin{tabular}{lccccccccc}
\toprule
Method & Prefill & KV & Single-Doc QA & Multi-Doc QA & Summarization & Few-shot & Synthetic & Code & Avg. \\
\midrule
\multicolumn{10}{c}{\textbf{LLaMA-3.1-8B-Instruct}} \\
\midrule
Full-context   & 100\% & 100\% & 43.58 & 44.65 & 29.22 & 69.48 & 54.21 & 60.01 & 50.19 \\
\midrule
\multicolumn{10}{c}{\textbf{Decoding-Only}} \\
\midrule
StreamingLLM   & 100\% & 10\% & 28.67 & 37.01 & 22.28 & 63.19 & 47.96 & 58.01 & 42.85 \\
               & 100\% & 20\% & 30.43 & 38.89 & 24.19 & 65.41 & 46.83 & 59.44 & 44.20 \\
H2O            & 100\% & 10\% & 27.12 & 19.74 & 25.96 & 66.65 & 34.58 & 54.88 & 38.15 \\
               & 100\% & 20\% & 30.66 & 19.69 & 26.05 & 68.00 & 36.28 & 58.05 & 39.79 \\
SnapKV         & 100\% & 10\% & 41.90 & 44.11 & 25.34 & 68.17 & 53.72 & 59.53 & 48.73 \\
               & 100\% & 20\% & 43.27 & 43.92 & 26.75 & 68.21 & 53.73 & 60.73 & 49.43 \\
\midrule
\multicolumn{10}{c}{\textbf{Prefill-Aware}} \\
\midrule
PyramidInfer   & 60\%  & 60\%  & 28.41 & 19.31 & 23.50 & 66.40 & 36.87 & 61.41 & 39.32 \\
GemFilter      & 51\%  & 10\%  & 28.39 & 40.33 & 21.78 & 64.31 & 46.13 & 30.75 & 38.61 \\
               & 61\%  & 20\%  & 34.63 & 40.77 & 24.39 & 66.39 & 51.59 & 36.56 & 42.39 \\
\rowcolor{blue!10}
FastKV         & 60\%  & 10\%  & 41.95 & 43.73 & 24.78 & 69.45 & 53.43 & 57.45 & 48.47 \\
\rowcolor{blue!10}
               & 60\%  & 20\%  & 42.93 & 43.74 & 26.03 & 69.51 & 53.56 & 68.64 & 49.07 \\
\midrule
\multicolumn{10}{c}{\textbf{Ministral-8B-Instruct}} \\
\midrule
Full-context   & 100\% & 100\% & 41.58 & 49.45 & 27.72 & 70.83 & 54.50 & 67.13 & 51.87 \\
\midrule
\multicolumn{10}{c}{\textbf{Decoding-Only}} \\
\midrule
StreamingLLM   & 100\% & 10\% & 27.74 & 35.87 & 21.00 & 65.44 & 34.50 & 62.70 & 41.21 \\
               & 100\% & 20\% & 29.14 & 38.89 & 23.23 & 68.01 & 38.50 & --    & --    \\
H2O            & 100\% & 10\% & 37.47 & 43.88 & 25.96 & 68.64 & 38.75 & 60.95 & 45.94 \\
               & 100\% & 20\% & 38.93 & 44.27 & 26.38 & 69.30 & 39.25 & 63.58 & 46.95 \\
SnapKV         & 100\% & 10\% & 39.97 & 49.43 & 23.92 & 70.03 & 55.00 & 65.75 & 50.68 \\
               & 100\% & 20\% & 40.63 & 49.59 & 25.67 & 70.64 & 54.50 & 67.07 & 51.35 \\
\midrule
\multicolumn{10}{c}{\textbf{Prefill-Aware}} \\
\midrule
PyramidInfer   & 60\%  & 60\%  & 34.29 & 37.34 & 24.16 & 68.19 & 52.25 & 65.83 & 47.01 \\
GemFilter      & 60\%  & 10\%  & 37.92 & 55.22 & 23.30 & 65.05 & 46.00 & 30.88 & 42.82 \\
               & 70\%  & 20\%  & 40.84 & 54.06 & 25.31 & 68.20 & 51.25 & 38.56 & 46.37 \\
\rowcolor{blue!10}
FastKV         & 60\%  & 10\%  & 39.71 & 49.63 & 23.38 & 70.71 & 55.00 & 64.00 & 50.40 \\
\rowcolor{blue!10}
               & 60\%  & 20\%  & 41.08 & 49.20 & 25.02 & 71.41 & 55.00 & 64.68 & 51.07 \\
\bottomrule
\end{tabular}
}
}
\end{table*}

\begin{table}[t]
  \centering
  \caption{RULER results on LLaMA-3.1-8B-Instruct.}
  \label{tab:ruler}
  \setlength{\tabcolsep}{1.8pt}
  \renewcommand{\arraystretch}{0.85}
  \scalebox{0.78}{
  \begin{tabular}{lcccccccc}
    \toprule
    \textbf{Method} & \textbf{Prefill} & \textbf{KV} & \textbf{8K} & \textbf{16K} & \textbf{32K} & \textbf{64K} & \textbf{128K} & \textbf{Avg.} \\
    \midrule
    Full-context   & 100\% & 100\% & 90.1 & 95.0 & 83.4 & 85.5 & 76.3 & 86.0 \\
    StreamingLLM   & 100\% & 10\%  & 15.0 & 21.5 & 24.4 & 16.9 & 15.0 & 18.6 \\
    H2O            & 100\% & 10\%  & 27.1 & -    & -    & - & - & - \\
    SnapKV         & 100\% & 10\%  & 75.6 & 76.8 & 72.9 & 75.0 & 67.7 & 73.6 \\
    PyramidInfer   & 60\%  & 60\%  & 66.5 & -    & -    & - & - & - \\
    GemFilter      & 51\%  & 10\%  & 69.7 & 68.2 & 70.4 & 69.8 & 69.8 & 69.6 \\
    \rowcolor{blue!10}
    FastKV         & 60\%  & 10\%  & 77.8 & 77.3 & 77.2 & 77.4 & 68.2 & 75.6 \\
    \bottomrule
  \end{tabular}
  }
    \vspace{-2mm}
\end{table}

\noindent\textbf{RULER.}
We present the RULER evaluation results for LLaMA-3.1-8B-Instruct with a KV retention rate of 10\% in Table~\ref{tab:ruler}. 
The evaluation is conducted with input context lengths up to 128K tokens.
Notably, FastKV outperforms other methods even under 128K inputs.
Taken together with the LongBench evaluation, these results demonstrate that FastKV effectively offers the best accuracy–latency trade-off across diverse long-context tasks while achieving efficient KV cache compression and acceleration for prefill and decoding stages.

\begin{table}[t]
  \centering
  \caption{Needle-in-a-Haystack results on LLaMA-3.1-8B-Instruct.}
  \label{tab:niah}
  \setlength{\tabcolsep}{14pt}
  \renewcommand{\arraystretch}{0.85}
  \scalebox{0.78}{
  \begin{tabular}{lccc}
    \toprule
    \textbf{Method} & \textbf{Prefill} & \textbf{KV} & \textbf{Score} \\
    \midrule
    Full-context   & 100\% & 100\% & 99.0 \\
    StreamingLLM   & 100\% & 10\%  & 33.5 \\
    SnapKV         & 100\% & 10\%  & 99.0 \\
    GemFilter      & 51\%  & 10\%  & 95.8 \\
    \rowcolor{blue!10}
    FastKV         & 60\%  & 10\%  & 99.9 \\
    \bottomrule
  \end{tabular}
  }
  \vspace{-4mm}
\end{table}

\noindent\textbf{Needle-in-a-Haystack.}
Table~\ref{tab:niah} shows the Needle-in-a-Haystack evaluations for LLaMA-3.1-8B-Instruct with a KV retention rate of 10\% budget. 
Score denotes the average accuracy across sequence lengths 16K, 32K, 48K, 64K, 80K, 96K, 112K, and 128K.
Results for each sequence length are provided in the Appendix~\ref{appendix:more_experimental_results_niah}.
Consistent with the results observed in LongBench and RULER benchmark, FastKV achieves the best performance.

\subsection{Latency}

\begin{figure*}[t]
    \centering
    \includegraphics[width=0.95\textwidth]{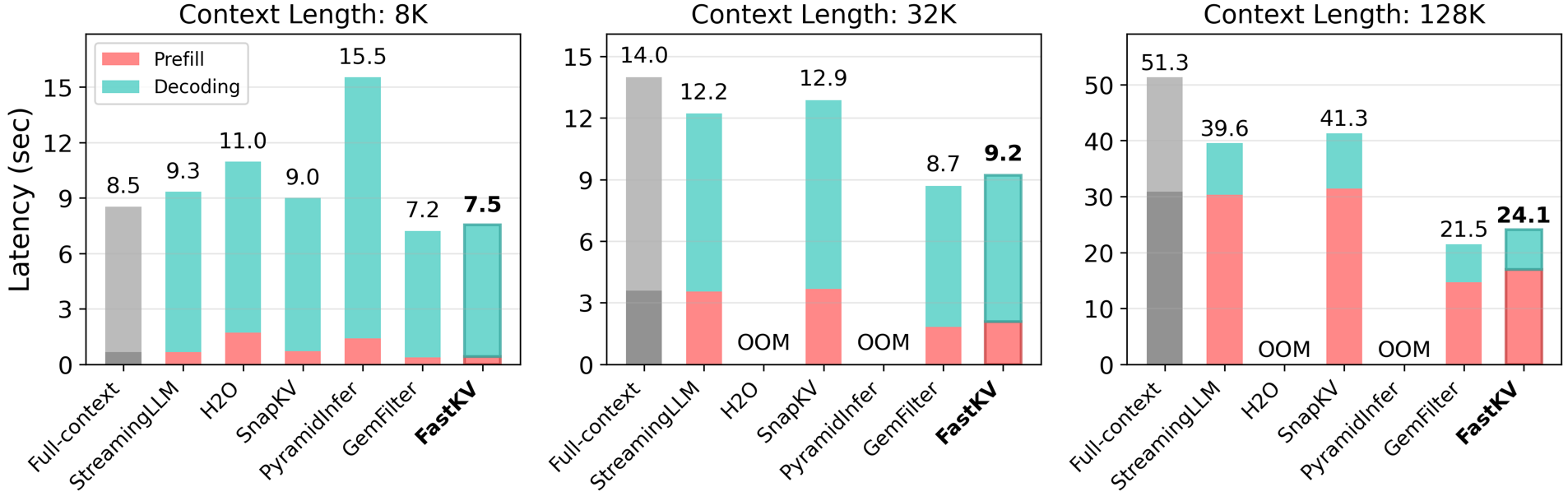}
    \vspace{-3mm}
    \caption{End-to-end inference latency breakdown of LLaMA-3.1-8B-Instruct at varying input context lengths (generating 256 tokens).}
    \label{fig4}
    \vspace{-4mm}
\end{figure*}

We benchmark end-to-end latency using LLaMA-3.1-8B-Instruct on a single A100 SXM GPU. 
Results for Ministral-8B-Instruct are reported in the Appendix~\ref{appendix:more_experimental_results}. 
All experiments fix the generation to 256 tokens, while the input context length varies.

As shown in Figure~\ref{fig4}, end-to-end latency increases rapidly with longer input context under full-context execution, where prefill latency dominates at 128K. 
Decoding-only approaches such as StreamingLLM and SnapKV reduce decoding time, but their benefit diminishes when prefill dominates; in particular, SnapKV stores KV states per attention head, which limits decoding speedup in GQA models.  
H2O, which requires exporting the entire attention map, cannot leverage FlashAttention-2 and is slower than full-context even at 8K, while running out of memory at longer contexts.  
Similarly, PyramidInfer does not employ FlashAttention-2 in its implementation and therefore causes an out-of-memory error at context length beyond 8K.
Even at 8K context length, it shows substantially long decoding latency suffering from large KV cache size due to its enforced high KV retention rate.

In contrast, both GemFilter and FastKV effectively reduce both prefill and decoding latency. 
Their advantage is most pronounced at 128K, where they achieve more than $2\times$ speedup over the full-context baseline. 
GemFilter’s slight advantage is primarily attributed to its low prefill compute rate, determined by the filter layer selection (13), compared to the TSP layer (15). However, as discussed in Section~\ref{subsec:accuracy}, GemFilter is more prone to accuracy degradation. 

\subsection{Ablation Studies}

We conduct ablation studies on LLaMA-3.1-8B-Instruct to examine the effect of the two key hyperparameters of FastKV: TSP rate and TSP layer index.

\begin{figure}[t]
    \centering
    \includegraphics[width=1.0\columnwidth]{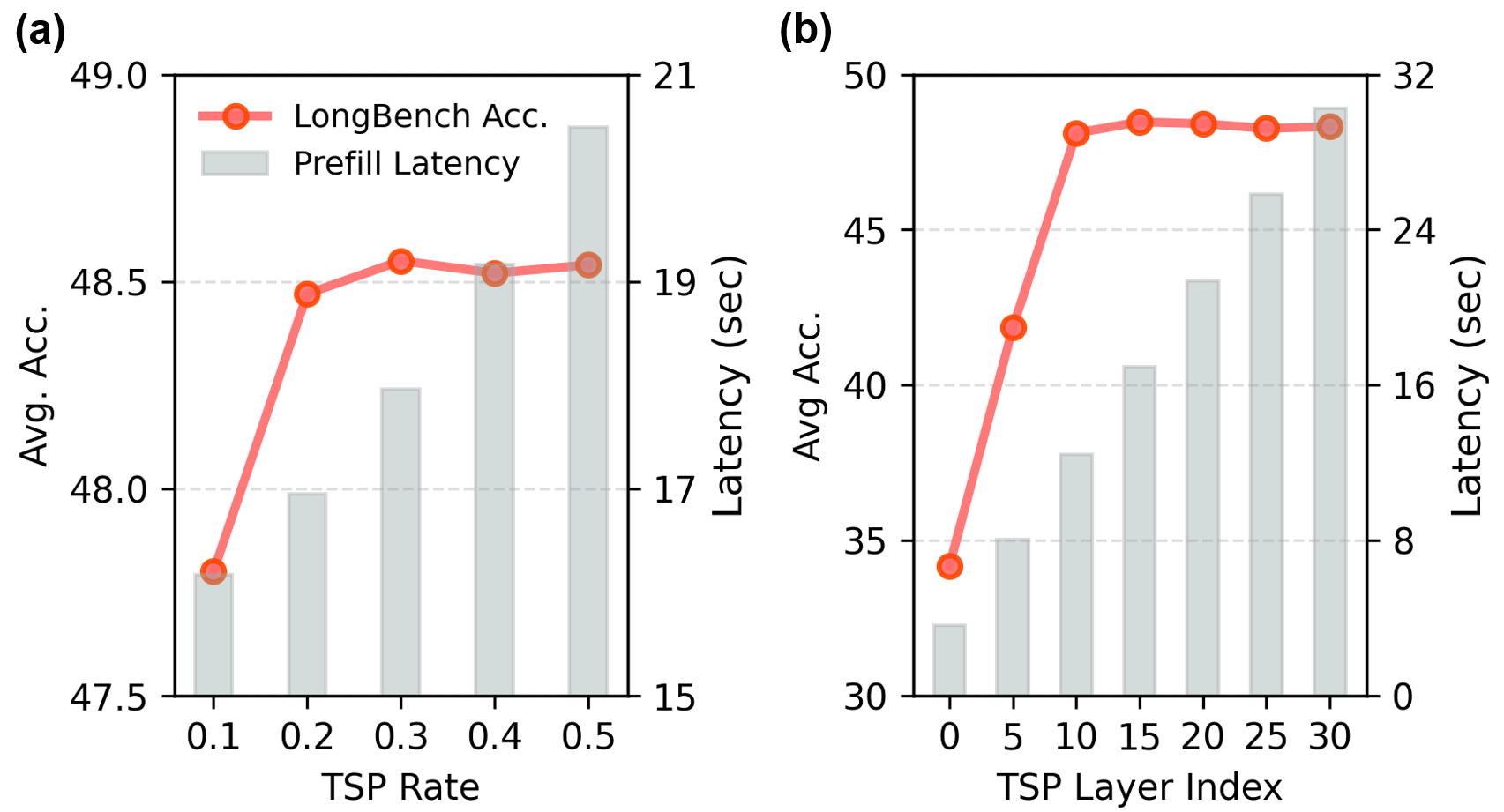}
    \vspace{-6mm}
    \caption{(a) Effect of TSP rate on LongBench average accuracy and prefill latency. (b) Effect of TSP layer index on LongBench average accuracy and prefill latency.}
    \label{fig5}
    \vspace{-4mm}
\end{figure}

\noindent \textbf{TSP Rate.}
Figure~\ref{fig5}(a) shows the effect of varying the TSP rate while fixing the TSP layer index and retention rate at 15 and 10\% respectively. 
When the rate is set too low, only a very small subset of tokens is propagated to later layers, which limits the model’s ability to convey sufficient context and leads to accuracy degradation. 
Increasing the rate improves accuracy by allowing more salient tokens to be propagated, and performance stabilizes around 20\%, which we adopt as the default setting.

\noindent \textbf{TSP Layer Index.}
Figure~\ref{fig5}(b) reports the impact of the TSP layer index under a fixed TSP rate of 20\% and retention rate of 10\%. 
Applying TSP at earlier layers leads to substantial accuracy degradation. 
This is because early layers exhibit highly dynamic attention patterns, where the set of critical tokens frequently shifts and discarding tokens too early removes information that later layers would still require. 
By contrast, later layers show more stable token dependencies, making selective propagation more reliable. 
The proposed selection algorithm identifies layer 15 as an optimal point where the LongBench score saturates, indicating a favorable trade-off between speed and accuracy.

\section{Conclusion}

In this work, we introduced FastKV, a framework that accelerates long-context inference through a two-stage prefill strategy incorporating Token-Selective Propagation and a KV cache compression that allocates KV retention rate independently of the prefill compute rate. 
Experiments show that FastKV achieves up to 1.82$\times$ faster prefill and 2.87$\times$ faster decoding while maintaining the accuracy of full-context baselines.
Beyond raw speedup, FastKV demonstrates that accounting for the dynamics of critical context across layers is crucial for efficient long-context processing. 
This design principle enables scalable inference as context lengths continue to grow, making long-context inference more practical. 

\section*{Limitations}

While FastKV yields substantial acceleration in both prefill and decoding stages, its advantage becomes more evident in long-context scenarios where KV cache handling dominates the overall latency.
For short-context inputs, however, the KV cache is not a major overhead, and alternative acceleration methods such as quantization~\citep{awq, gptq, quarot} or optimized kernels~\citep{flashattention3, flashinfer} may be more beneficial. 
Combining FastKV with these complementary techniques could enable more consistent efficiency across diverse inference settings.
Beyond these research directions, FastKV is readily compatible with modern serving frameworks~\citep{vllm, sglang}. It is orthogonal to batching and paged attention.
These directions highlight promising opportunities for extending FastKV toward more adaptive and universally efficient LLM inference.

\section*{Ethical Considerations}

This work focuses on improving inference efficiency of large language models. FastKV reduces computational and memory overhead with minimal impact on output behavior. 
The method operates at the architectural level and does not introduce new ethical or societal risks beyond those already associated with large language models.

\section*{Acknowledgements}
This work was supported in part by Institute of Information $\&$ communications Technology Planning $\&$ Evaluation (IITP) grant funded by the Korea government (MSIT) (No.RS-2025-02273157: Development of Low Power Training/Inference Technologies based on AI Semiconductors, RS-2023-00256081: artificial intelligence semiconductor support program to nurture the best talents), AI Computing Infrastructure Enhancement (GPU Rental Support) User Support Program funded by MSIT(No.RQT-25-090058), Samsung Research Funding Center under Project SRFC-TC1603-53, and BK21 FOUR program. (Corresponding Authors: Yulhwa Kim and Jae-Joon Kim).

\bibliography{custom}

\onecolumn
\appendix

\section{Related Works}

Research efforts to alleviate the burden of long-context inference in LLMs have explored multiple directions.
One body of work is sparse attention~\citep{minference,flexprefill,peng2025accelerating} which reduces the amount of computation required for generating the KV cache.
They focus on identifying inherent sparse patterns in attention map and leverage them to design optimized kernels.

Methods that compress the generated KV cache are typically categorized into quantization-based and pruning-based approaches. 
Quantization-based methods, such as KVQuant~\citep{kvquant}, KIVI~\citep{kivi} and MiKV~\citep{mikv}, focus on reducing the precision of key and value tensors to lower memory and bandwidth requirements without altering the cache structure itself. While these techniques effectively reduce storage and transfer costs, they still do not alleviate the quadratic computation in the prefill stage, limiting their impact on overall latency.
Pruning-based methods~\citep{streamingllm,h2o,snapkv} include the baselines of this work, dropping KV cache in token-wise manner.
These approaches directly bring speed up in decoding stage without complex kernel-level optimizations since overhead of loading KV cache and computation itself is reduced.
Some works~\citep{pyramidinfer, gemfilter} aim to reduce the number of hidden states processed during the prefill stage, thereby generating smaller KV cache and achieving acceleration in both prefill and decoding stages.

FastKV approaches the problem from the perspective of KV cache pruning, aiming to achieve prefill compute reduction as well.
It is orthogonal to kernel-level sparse attention and KV cache quantization techniques, and thus can be combined with them to further extend efficiency.

\section{Additional Methodological Details}
\label{appendix:methodological_details}
\subsection{FastKV Algorithm}
We present the pseudocode for the FastKV algorithm, which performs two complementary compression strategies during the prefill stage: (1) reducing hidden states via Token-Selective Propagation (TSP), and (2) compressing the attention key-value (KV) cache at each layer. 
These two steps are carried out independently: TSP selects a subset of hidden states to propagate to later layers, while KV cache compression prunes KV cache entries at each layer using attention-based importance scores. 
Together, they improve prefill efficiency and reduce memory usage without modifying the model architecture or introducing any runtime overhead during decoding.

\vspace{2mm}

\noindent The key terms used in the pseudocode are:
\begin{itemize}
\item \textbf{\textit{L\textsubscript{TSP}}}: Index of TSP layer

\item \textbf{\textit{R\textsubscript{TSP}}}: TSP rate

\item \textbf{\textit{R\textsubscript{KV}}}: KV retention rate

\item \textbf{\textit{KVCompress}}: Selects the KV entries of top-\{context length $\times$ \textbf{\textit{R\textsubscript{KV}}}\} critical tokens based on group-wise saliency scores.
The group-wise scores are obtained by averaging head-wise saliency values computed using Equation~\ref{eq:TSP-1} within each key-value group.
\item \textbf{\textit{HiddenCompress}}: Selects hidden states of top-\{context length $\times$ \textbf{\textit{R\textsubscript{TSP}}}\} critical tokens, as determined by Equation~\ref{eq:TSP-2}, to propagate to next layer.

\end{itemize}

\begin{algorithm*}
\caption{FastKV algorithm for the KV cache compression during the prefill stage}
\begin{algorithmic}[1]
\Require $input$ $sequence$ $\{I\},$ $\#layers$ $\{L\},$ $L_{TSP},$ $R_{TSP},$ $R_{KV}$
\Ensure $generated$ $token$ $\{O\},$ $KV$ $Cache$ $\{C\}$
\State $X \leftarrow Embedding(I)$ 
    \For{$l$ = 0 {\bfseries to} $L-1$}
        \If{$l \leq L_{TSP}$}
            \State $X,Att_l,K_X,V_X\leftarrow layer_l(X)$ 
            \Comment{$X:$  $hidden$  $states$  $before$  $TSP$}
            \State $K,V\leftarrow KVCompress(K_X,V_X,Att_l,R_{KV})$ 
            \If{$l==L_{TSP}$}
                \State $x \leftarrow HiddenCompress(X,Att_l,R_{TSP})$
                \Comment{$x:$  $reduced$  $hidden$  $states$}
            \EndIf
        \Else
            \State $x,Att_l,K_x,V_x=layer_l(x)$
            \State $K,V\leftarrow KVCompress(K_x,V_x,Att_l,R_{KV})$
        \EndIf
        \State $C\leftarrow update(K,V)$
    \EndFor
    \State $O\leftarrow LMHead(x)$
    \State \Return{$O,C$}

\end{algorithmic}
\label{algo:fastkv}
\end{algorithm*}

\newpage
\subsection{Distinction from Prefill-Aware KV Cache Compression Methods}
\label{appendix:distinction}

\begin{figure*}[h]
\centering
\includegraphics[width=1.0\linewidth]{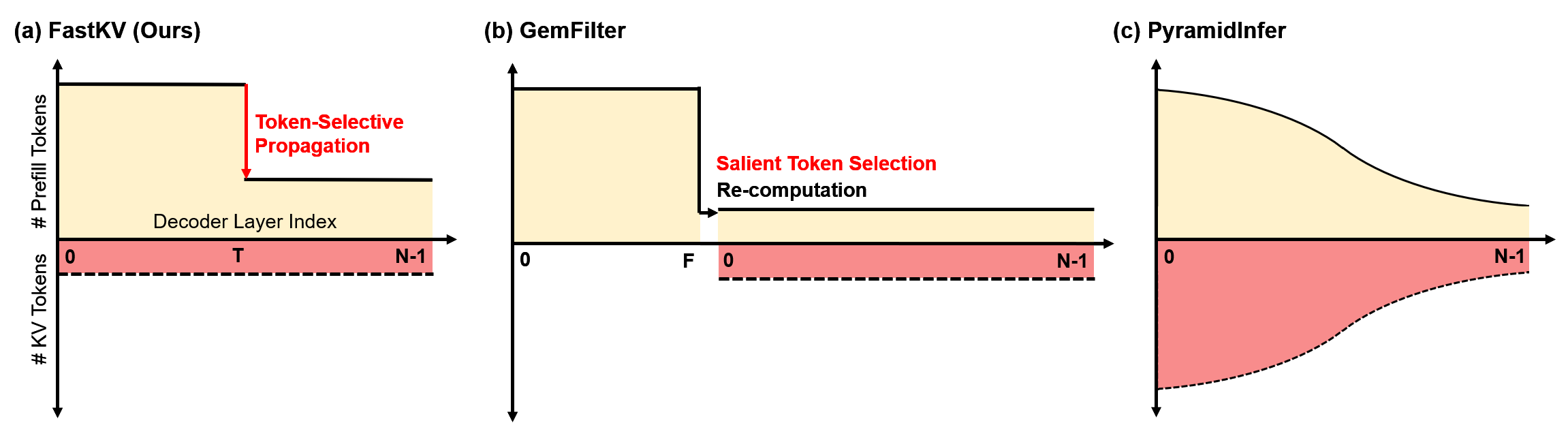}
\caption{
Comparison of processing flows between GemFilter and FastKV. GemFilter prunes the input prompt and recomputes later layers with only selected tokens, while FastKV propagates the full-context to a mid-layer before reducing the token set for further computation.
}
\label{fig6}
\end{figure*}

We compare FastKV with baseline prefill-aware KV cache compression methods.
Figure~\ref{fig6} illustrates the differences among the three approaches.

First, FastKV propagates the hidden states of all input context tokens up to the Token-Selective Propagation (TSP) layer $T$. 
At this layer, Token-Selective Propagation occurs: only the selected salient tokens and window tokens are passed to the later layers. 
Consequently, later layers process fewer tokens than the early layers, since they operate on the reduced token set. 
Meanwhile, KV cache compression is applied independently at each decoder layer once its prefill computation is complete, according to a separately configured KV retention rate.

In contrast, GemFilter performs prefill up to the filter layer $F$ in order to select salient tokens. 
Once this set is fixed, the prefill stage is restarted with the reduced token set. 
This policy enforces the salient token selection made at the filter layer across all decoder layers, which can be problematic for early layers where attention score distributions vary significantly. 
Moreover, the prefill compute rate is directly tied to the KV retention rate, preventing independent control of prefill compute and KV budget.

PyramidInfer, on the other hand, gradually reduces the number of tokens processed at each decoder layer during prefill. 
The decay rate follows a cosine schedule, which can lead to premature context reduction before the critical context is stabilized. In addition, the full KV cache of all prefilled tokens in each layer is retained during decoding. 
As a result, in PyramidInfer the prefill compute rate is always equal to the KV retention rate, further restricting the ability to optimize the trade-off between decoding latency and accuracy.

\begin{figure*}[h]
\centering
\includegraphics[width=0.85\linewidth]{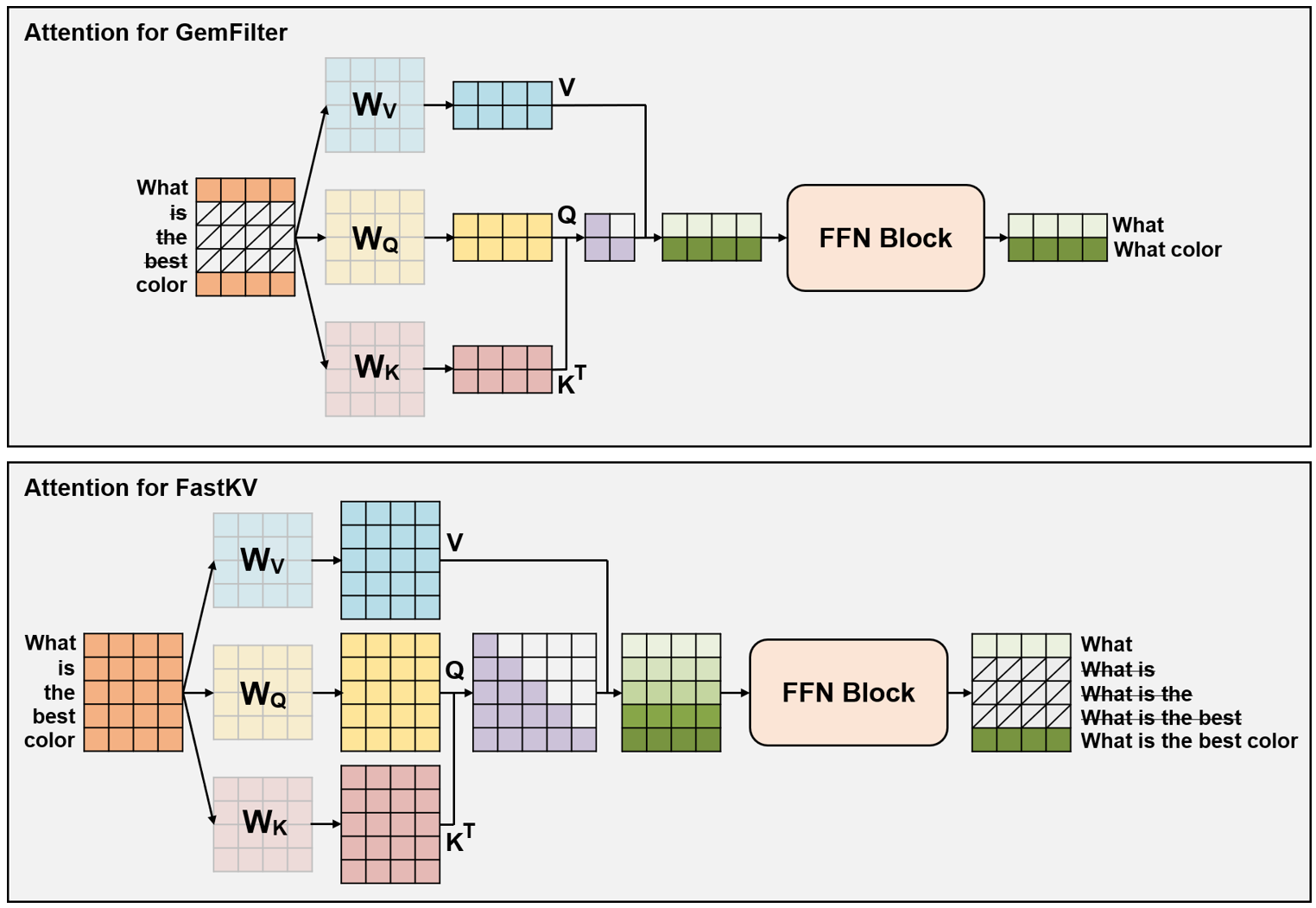}
\caption{
Visualization of attention computation. In GemFilter, discarded tokens are entirely excluded from attention, leading to fragmented context representations. FastKV discards tokens after the TSP layer, but their information has already been integrated through early-layer attention, allowing global semantics to be preserved despite later compression.
}
\label{fig7}
\end{figure*}

An additional limitation of GemFilter stems from its strategy of discarding non-salient tokens and restarting prefill.
As shown in Figure~\ref{fig7}, the information carried by discarded tokens is never represented, often leading to fragmented context understanding.
In contrast, FastKV allows tokens that are ultimately discarded by TSP to still contribute through attention computations in the early layers, ensuring that even these tokens can influence the output.
As a result, the later layers operating on the reduced hidden states are able to access semantically enriched representations.
This design enables FastKV to preserve contextual fidelity while still achieving substantial latency and memory savings.

\vspace{4pt}

\newpage
\section{Evaluation Details}
\label{appendix:evaluation_details}

\subsection{Models}
\label{subsec:models_details}

We conduct evaluations using three open-source LLMs which support 128K context window: LLaMA-3.1-8B-Instruct~\cite{llama3}, Ministral-8B-Instruct~\cite{ministral} and Mistral-Nemo-12B-Instruct~\cite{mistralnemo}.
LLaMA-3.1-8B-Instruct has 32 decoder layers and we choose layer 15 for the TSP layer.
Ministral-8B-Instruct has 36 decoder layers and we choose layer 17 for the TSP layer.
Mistral-Nemo-12B-Instruct has 40 decoder layers and we choose layer 19 for the TSP layer.

\sloppy

The checkpoints for all models are publicly available at:

\begin{tabularx}{\textwidth}{lX}
\textbf{LLaMA-3.1-8B-Instruct}: \\
\url{https://huggingface.co/meta-llama/Llama-3.1-8B-Instruct} \\
\textbf{Ministral-8B-Instruct}:\\
\url{https://huggingface.co/mistralai/Ministral-8B-Instruct-2410} \\
\textbf{Mistral-Nemo-12B-Instruct}: \\
\url{https://huggingface.co/nvidia/Mistral-NeMo-12B-Instruct}
\end{tabularx}

\newpage
\subsection{Datasets}
\label{subsec:dataset_details}

\noindent \textbf{LongBench.} 
LongBench is a collection of benchmark datasets designed to assess the long-context understanding capabilities of LLMs.
It includes tasks in both English and Chinese, spanning six categories, each consisting of multiple subtasks.

In this study, we evaluate FastKV on a subset of LongBench that includes only the English-language tasks and code-related benchmarks.
Detailed information for each selected subtask is provided in Table~\ref{tab:longbench_details}.

\begin{table*}[h]
\centering
\caption{LongBench dataset details.}
\label{tab:longbench_details}
\setlength{\tabcolsep}{5pt}
\renewcommand{\arraystretch}{1.0}
\scalebox{0.9}{
\begin{tabular}{lllccc}
\toprule
\textbf{Task Type} & \textbf{Task} & \textbf{Abbreviation} & \textbf{Eval metric} & \textbf{Language} & \textbf{\#Sample} \\
\midrule
\multirow{4}{*}{Single-doc QA} 
  & NarrativeQA & NrtvQA & F1 & EN & 200 \\
  & Qasper & Qasper & F1 & EN & 200 \\
  & MultiFieldQA-en & MF-en & F1 & EN & 150 \\
  & HotpotQA & HotpotQA & F1 & EN & 200 \\
\midrule
\multirow{2}{*}{Multi-doc QA}
  & 2WikiMultihopQA & 2WikiMultihopQA & F1 & EN & 200 \\
  & MuSiQue & MuSiQue & F1 & EN & 200 \\
\midrule
\multirow{3}{*}{Summarization}
  & GovReport & GovReport & Rouge-L & EN & 200 \\
  & QMSum & QMSum & Rouge-L & EN & 200 \\
  & MultiNews & MultiNews & Rouge-L & EN & 200 \\
\midrule
\multirow{3}{*}{Few shot}
  & TREC & TREC & Accuracy & EN & 200 \\
  & TriviaQA & TriviaQA & F1 & EN & 200 \\
  & SAMSum & SAMSum & Rouge-L & EN & 200 \\
\midrule
\multirow{2}{*}{Code}
  & LCC & LCC & Edit Sim & Python/C\#/Java & 500 \\
  & RepoBench-P & RB-P & Edit Sim & Python/Java & 500 \\
\midrule
\multirow{2}{*}{Synthetic}
  & PassageCount & PCount & Accuracy & EN & 200 \\
  & PassageRetrieval-en & PRe & Accuracy & EN & 200 \\
\bottomrule
\end{tabular}
}
\end{table*}

\noindent \textbf{RULER.}
RULER is a synthetic benchmark for evaluating long-context LLMs with flexible configurations for customized sequence length and task complexity. 
It extends the Needle-in-a-Haystack paradigm into a broader benchmark suite with 13 tasks across four categories: retrieval (NIAH-style), aggregation (CWE, FWE), multi-hop tracing (VT), and QA. 
This design enables systematic stress tests that probe retrieval, aggregation, and reasoning beyond simple lookup. 

In our evaluation, we used 11 benchmarks spanning retrieval, aggregation, and multi-hop tracing, excluding the two QA benchmarks.

\noindent\textbf{Needle-in-a-Haystack.}
Needle-in-a-Haystack is a benchmark designed to evaluate the in-context retrieval capabilities of LLMs under long-context settings.
A statement, referred to as the needle, is randomly selected and inserted at a specific position within a long input context.
The model is tasked with retrieving the needle content given the entire context.
Evaluation is conducted across various context lengths and needle depths (insertion positions) to systematically measure retrieval performance under different levels of difficulty.

In our experiments, retrieval performance is evaluated at context lengths ranging from 16K to 128K tokens, with measurements taken at 16K-token intervals.

\newpage
\section{More Experimental Results}
\label{appendix:more_experimental_results}

\subsection{LongBench}
\label{appendix:more_experimental_results_longbench}
We provide full breakdown of LongBench results for LLaMA-3.1-8B-Instruct and Ministral-8B-Instruct.
Table~\ref{tab:appendix_longbench} presents the score of each subtask of LongBench.

For H2O and PyramidInfer, we truncate inputs longer than 8K tokens to 8K tokens, as longer sequences cause out-of-memory errors due to their naive attention implementations, which are not optimized for long-context processing. 
We extended a common practice in LongBench evaluations when handling inputs that exceed a model’s supported context window size.

\newcolumntype{W}{>{\centering\arraybackslash}p{1.05cm}}

\begin{table*}[ht]
  \centering
  \setlength{\tabcolsep}{1.2pt}
  \caption{Detailed LongBench results for LLaMA-3.1-8B-Instruct and Ministral-8B-Instruct.}
  \label{tab:appendix_longbench}
  \renewcommand{\arraystretch}{0.95}
  \scalebox{0.7}{
    \begin{tabular}{lcc*{3}{W}*{3}{W}*{3}{W}*{3}{W}*{2}{W}*{2}{W}}
    \toprule
     & Prefill & KV 
     & \multicolumn{3}{c}{Single-Document QA}
     & \multicolumn{3}{c}{Multi-Document QA}
     & \multicolumn{3}{c}{Summarization}
     & \multicolumn{3}{c}{Few-shot Learning}
     & \multicolumn{2}{c}{Synthetic}
     & \multicolumn{2}{c}{Code} \\
    \cmidrule(lr){4-6}
    \cmidrule(lr){7-9}
    \cmidrule(lr){10-12}
    \cmidrule(lr){13-15}
    \cmidrule(lr){16-17}
    \cmidrule(lr){18-19}
    Method 
    & \rotatebox{60}{\makebox[1.3cm][c]{Compute}}
    & \rotatebox{60}{\makebox[1.3cm][c]{Retain}}
     & \rotatebox{60}{\makebox[1.3cm][c]{NrtvQA}} 
     & \rotatebox{60}{\makebox[1.3cm][c]{Qasper}}
     & \rotatebox{60}{\makebox[1.3cm][c]{MF-en}}
     & \rotatebox{60}{\makebox[1.3cm][c]{HotpotQA}}
     & \rotatebox{60}{\makebox[1.3cm][c]{2WikiMQA}}
     & \rotatebox{60}{\makebox[1.3cm][c]{Musique}}
     & \rotatebox{60}{\makebox[1.3cm][c]{GovReport}}
     & \rotatebox{60}{\makebox[1.3cm][c]{QMSum}}
     & \rotatebox{60}{\makebox[1.3cm][c]{MultiNews}}
     & \rotatebox{60}{\makebox[1.3cm][c]{TREC}}
     & \rotatebox{60}{\makebox[1.3cm][c]{TriviaQA}}
     & \rotatebox{60}{\makebox[1.3cm][c]{SAMSum}}
     & \rotatebox{60}{\makebox[1.3cm][c]{Pcount}}
     & \rotatebox{60}{\makebox[1.3cm][c]{PRe}}
     & \rotatebox{60}{\makebox[1.3cm][c]{LCC}}
     & \rotatebox{60}{\makebox[1.3cm][c]{RB-P}} \\
    \midrule
      \midrule
      \multicolumn{19}{c}{\textbf{LLaMA-3.1-8B-Instruct}} \\
      \midrule
      Full-context & 100\% & 100\% & 30.21 & 45.53 & 55.01 & 56.01 & 46.65 & 31.28 & 35.13 & 25.28 & 27.25 & 73.00 & 91.64 & 43.80 & 8.91 & 99.50 & 63.38 & 56.64 \\
      \midrule
      \multicolumn{19}{c}{\textbf{Decoding-Only}} \\
      \midrule
      StreamingLLM & 100\% & 10\% & 26.05 & 26.29 & 33.68 & 48.07 & 37.93 & 25.03 & 25.37 & 21.49 & 19.97 & 58.50 & 88.46 & 42.61 & 7.91 & 88.00 & 61.07 & 54.94 \\
                   & 100\% & 20\% & 27.99 & 28.89 & 34.42 & 51.21 & 41.03 & 24.43 & 28.34 & 22.08 & 22.16 & 64.00 & 89.96 & 42.27 & 8.66 & 85.00 & 62.02 & 56.85 \\
      H2O          & 100\% & 10\% & 9.32  & 34.96 & 37.08 & 19.65 & 32.65 & 6.91  & 31.29 & 21.01 & 25.59 & 66.50 & 91.03 & 42.16 & 3.41 & 65.75 & 58.27 & 51.48 \\
                   & 100\% & 20\% & 9.80  & 41.09 & 41.00 & 19.20 & 33.92 & 6.83  & 32.26 & 21.50 & 25.62 & 66.00 & 91.27 & 42.41 & 4.31 & 68.25 & 62.02 & 54.08 \\
      SnapKV       & 100\% & 10\% & 31.51 & 40.18 & 41.00 & 55.83 & 44.30 & 30.92 & 28.88 & 24.58 & 22.57 & 70.50 & 91.28 & 42.64 & 7.93 & 99.50 & 62.32 & 56.71 \\
                   & 100\% & 20\% & 31.15 & 44.11 & 54.55 & 55.47 & 45.14 & 31.14 & 31.22 & 24.94 & 24.09 & 70.50 & 91.90 & 42.23 & 7.96 & 99.50 & 63.55 & 57.91 \\
      \midrule
      \multicolumn{19}{c}{\textbf{Prefill-Aware}} \\
      \midrule
      PyramidInfer & 60\% & 60\% & 12.10 & 35.54 & 37.58 & 17.66 & 33.71 & 6.56  & 27.84 & 21.62 & 21.04 & 64.00 & 91.89 & 43.31 & 3.65 & 70.08 & 64.92 & 57.90 \\
      GemFilter    & 51\% & 10\% & 24.36 & 21.07 & 39.73 & 25.99 & 43.92 & 35.78 & 28.94 & 21.42 & 17.62 & 61.00 & 91.53 & 40.88 & 4.76 & 85.00 & 59.95 & 44.38 \\
                   & 61\% & 20\% & 26.02 & 30.74 & 47.30 & 37.97 & 55.97 & 20.64 & 31.19 & 20.92 & 21.06 & 61.50 & 92.75 & 40.92 & 6.18 & 97.00 & 31.99 & 41.12 \\
      \rowcolor{blue!5}
      FastKV       & 60\% & 10\% & 30.54 & 40.75 & 54.57 & 54.33 & 46.30 & 30.55 & 28.15 & 24.29 & 21.91 & 73.00 & 92.38 & 42.96 & 7.36 & 99.50 & 60.11 & 54.79 \\
      \rowcolor{blue!5}
                   & 60\% & 20\% & 30.26 & 43.70 & 54.83 & 54.39 & 46.42 & 30.42 & 30.28 & 24.88 & 22.93 & 73.50 & 91.97 & 43.07 & 7.61 & 99.50 & 61.61 & 55.67 \\
      \midrule
      \midrule
      \multicolumn{19}{c}{\textbf{Ministral-8B-Instruct}} \\
      \midrule
      Full-context & 100\% & 100\% & 24.55 & 47.99 & 52.21 & 60.43 & 52.64 & 35.28 & 32.36 & 24.16 & 26.64 & 74.50 & 92.04 & 45.94 & 9.00 & 100.00 & 67.06 & 67.19 \\
      \midrule
      \multicolumn{19}{c}{\textbf{Decoding-Only}} \\
      \midrule
      StreamingLLM & 100\% & 10\% & 22.52 & 33.02 & 27.69 & 48.68 & 37.66 & 21.26 & 23.55 & 21.09 & 18.36 & 61.50 & 91.38 & 43.45 & 6.00 & 63.00 & 62.84 & 62.55 \\
                   & 100\% & 20\% & 23.25 & 35.91 & 28.26 & 51.46 & 41.13 & 24.08 & 26.88 & 21.38 & 21.43 & 67.50 & 91.88 & 44.65 & 5.00 & 72.00 & 63.17 & 63.21 \\
      H2O          & 100\% & 10\% & 27.75 & 40.61 & 44.04 & 52.19 & 48.78 & 30.68 & 29.40 & 23.14 & 25.35 & 72.00 & 90.66 & 43.26 & 7.50 & 70.00 & 61.99 & 59.91 \\
                   & 100\% & 20\% & 27.28 & 43.31 & 44.57 & 53.41 & 49.93 & 30.88 & 30.47 & 23.55 & 25.87 & 72.00 & 91.16 & 44.04 & 7.50 & 71.00 & 65.02 & 66.47 \\
      SnapKV       & 100\% & 10\% & 26.82 & 43.55 & 49.53 & 60.18 & 51.72 & 36.38 & 26.94 & 23.55 & 21.27 & 72.00 & 92.04 & 44.01 & 10.00 & 100.00 & 65.02 & 66.47 \\
                   & 100\% & 20\% & 26.72 & 45.34 & 49.83 & 61.11 & 52.07 & 36.58 & 29.36 & 23.99 & 23.65 & 74.50 & 92.04 & 45.37 & 9.00 & 100.00 & 65.55 & 65.65 \\
      \midrule
      \multicolumn{19}{c}{\textbf{Prefill-Aware}} \\
      \midrule
      PyramidInfer & 60\% & 60\% & 24.92 & 37.60 & 40.34 & 48.92 & 40.84 & 22.26 & 27.23 & 22.99 & 22.25 & 68.50 & 91.06 & 45.02 & 7.50 & 97.00 & 66.91 & 64.74 \\
      GemFilter    & 60\% & 10\% & 25.70 & 37.97 & 39.73 & 60.62 & 56.04 & 37.22 & 28.65 & 21.18 & 19.57 & 63.00 & 89.36 & 42.79 & 6.50 & 85.50 & 59.96 & 41.79 \\
                   & 70\% & 20\% & 28.06 & 43.81 & 50.56 & 63.46 & 59.43 & 39.25 & 30.59 & 23.31 & 22.03 & 70.50 & 90.84 & 43.26 & 5.50 & 97.00 & 27.56 & 46.03 \\
      \rowcolor{blue!5}
      FastKV       & 60\% & 10\% & 25.31 & 43.71 & 50.11 & 61.31 & 50.85 & 36.73 & 26.15 & 23.45 & 20.54 & 75.00 & 92.04 & 45.10 & 10.00 & 100.00 & 62.99 & 65.00 \\
      \rowcolor{blue!5}
                   & 60\% & 20\% & 25.90 & 46.05 & 51.29 & 61.24 & 50.21 & 36.15 & 28.65 & 23.84 & 22.57 & 76.00 & 92.04 & 46.20 & 10.00 & 100.00 & 64.68 & 64.68 \\
      \bottomrule
    \end{tabular}
  }
  
\end{table*}

For a more comprehensive evaluation, Table~\ref{tab:appendix_longbench_nemo} additionally reports LongBench results on Mistral-Nemo-12B-Instruct, which is larger than the other two 8B-class models. 
On this model, FastKV also achieves stronger performance than the other baselines.

\begin{table*}[ht]
  \centering
  \setlength{\tabcolsep}{1.2pt}
  \caption{Detailed LongBench results for Mistral-Nemo-12B-Instruct.}
  \label{tab:appendix_longbench_nemo}
  \renewcommand{\arraystretch}{0.95}
  \scalebox{0.7}{
    \begin{tabular}{lcc*{3}{W}*{3}{W}*{3}{W}*{3}{W}*{2}{W}*{2}{W}}
    \toprule
     & Prefill & KV 
     & \multicolumn{3}{c}{Single-Document QA}
     & \multicolumn{3}{c}{Multi-Document QA}
     & \multicolumn{3}{c}{Summarization}
     & \multicolumn{3}{c}{Few-shot Learning}
     & \multicolumn{2}{c}{Synthetic}
     & \multicolumn{2}{c}{Code} \\
    \cmidrule(lr){4-6}
    \cmidrule(lr){7-9}
    \cmidrule(lr){10-12}
    \cmidrule(lr){13-15}
    \cmidrule(lr){16-17}
    \cmidrule(lr){18-19}
    Method 
    & \rotatebox{60}{\makebox[1.3cm][c]{Compute}}
    & \rotatebox{60}{\makebox[1.3cm][c]{Retain}}
     & \rotatebox{60}{\makebox[1.3cm][c]{NrtvQA}} 
     & \rotatebox{60}{\makebox[1.3cm][c]{Qasper}}
     & \rotatebox{60}{\makebox[1.3cm][c]{MF-en}}
     & \rotatebox{60}{\makebox[1.3cm][c]{HotpotQA}}
     & \rotatebox{60}{\makebox[1.3cm][c]{2WikiMQA}}
     & \rotatebox{60}{\makebox[1.3cm][c]{Musique}}
     & \rotatebox{60}{\makebox[1.3cm][c]{GovReport}}
     & \rotatebox{60}{\makebox[1.3cm][c]{QMSum}}
     & \rotatebox{60}{\makebox[1.3cm][c]{MultiNews}}
     & \rotatebox{60}{\makebox[1.3cm][c]{TREC}}
     & \rotatebox{60}{\makebox[1.3cm][c]{TriviaQA}}
     & \rotatebox{60}{\makebox[1.3cm][c]{SAMSum}}
     & \rotatebox{60}{\makebox[1.3cm][c]{Pcount}}
     & \rotatebox{60}{\makebox[1.3cm][c]{PRe}}
     & \rotatebox{60}{\makebox[1.3cm][c]{LCC}}
     & \rotatebox{60}{\makebox[1.3cm][c]{RB-P}} \\
    \midrule
      \midrule
      \multicolumn{19}{c}{\textbf{Mistral-Nemo-12B-Instruct}} \\
      \midrule
      Full-context & 100\% & 100\% & 26.27 & 43.64 & 58.11 & 49.34 & 45.85 & 26.26 & 31.31 & 24.15 & 26.08 & 75.00 & 89.66 & 44.32 & 1.50 & 98.00 & 68.58 & 68.11 \\
      \midrule
      \multicolumn{19}{c}{\textbf{Decoding-Only}} \\
      \midrule
      StreamingLLM & 100\% & 10\% & 21.50 & 27.54 & 31.03 & 35.91 & 37.13 & 18.21 & 18.45 & 16.11 & 17.76 & 63.00 & 89.88 & 35.49 & 1.50 & 55.00 & 63.17 & 57.81 \\
                   & 100\% & 20\% & 22.41 & 29.18 & 33.31 & 39.43 & 39.06 & 18.91 & 21.65 & 18.46 & 21.35 & 68.00 & 89.43 & 36.49 & 1.50 & 65.00 & 65.62 & 61.59 \\
      H2O          & 100\% & 10\% & 29.04 & 37.39 & 42.47 & 47.32 & 45.24 & 22.56 & 27.75 & 21.06 & 24.23 & 70.50 & 90.16 & 43.40 & 1.00 & 70.00 & 65.59 & 61.40 \\
                   & 100\% & 20\% & 30.48 & 38.46 & 49.63 & 49.23 & 46.57 & 22.57 & 29.56 & 22.19 & 25.02 & 71.50 & 90.59 & 43.69 & 3.50 & 70.50 & 67.58 & 63.02 \\
      SnapKV       & 100\% & 10\% & 24.96 & 39.68 & 56.36 & 49.15 & 45.27 & 26.18 & 25.97 & 23.33 & 21.60 & 73.50 & 89.66 & 43.54 & 1.50 & 97.50 & 66.93 & 65.99 \\
                   & 100\% & 20\% & 25.72 & 40.82 & 55.68 & 49.09 & 46.56 & 26.54 & 28.25 & 24.36 & 23.46 & 74.00 & 89.66 & 44.27 & 1.50 & 97.50 & 67.81 & 67.45 \\
      \midrule
      \multicolumn{19}{c}{\textbf{Prefill-Aware}} \\
      \midrule
      PyramidInfer & 60\% & 60\% & 30.45 & 33.99 & 44.59 & 45.99 & 43.53 & 22.31 & 27.46 & 21.43 & 23.75 & 68.50 & 89.74 & 43.74 & 1.00 & 71.00 & 69.57 & 66.73 \\
      GemFilter    & 60\% & 10\% & 33.33 & 36.81 & 52.02 & 55.39 & 57.08 & 37.36 & 27.92 & 20.46 & 20.60 & 64.50 & 86.39 & 42.15 & 3.00 & 91.00 & 33.78 & 45.00 \\
                   & 70\% & 20\% & 28.15 & 40.07 & 54.34 & 57.09 & 53.88 & 33.33 & 29.89 & 22.40 & 22.74 & 68.00 & 88.67 & 43.26 & 2.50 & 98.50 & 38.74 & 47.15 \\
      \rowcolor{blue!5}
      FastKV       & 60\% & 10\% & 25.97 & 40.14 & 57.89 & 49.84 & 46.47 & 24.94 & 25.37 & 22.73 & 20.73 & 74.50 & 90.06 & 44.48 & 0.50 & 99.00 & 66.43 & 65.40 \\
      \rowcolor{blue!5}
                   & 60\% & 20\% & 26.98 & 41.69 & 56.93 & 49.71 & 47.26 & 25.78 & 27.78 & 23.71 & 22.93 & 75.50 & 90.06 & 44.70 & 0.50 & 98.00 & 67.72 & 66.59 \\
    \bottomrule

    \end{tabular}
  }
  
\end{table*}

\newpage
\subsection{Needle-in-a-Haystack}
\label{appendix:more_experimental_results_niah}

\begin{figure*}[h]
    \centering
    \includegraphics[width=0.95\textwidth]{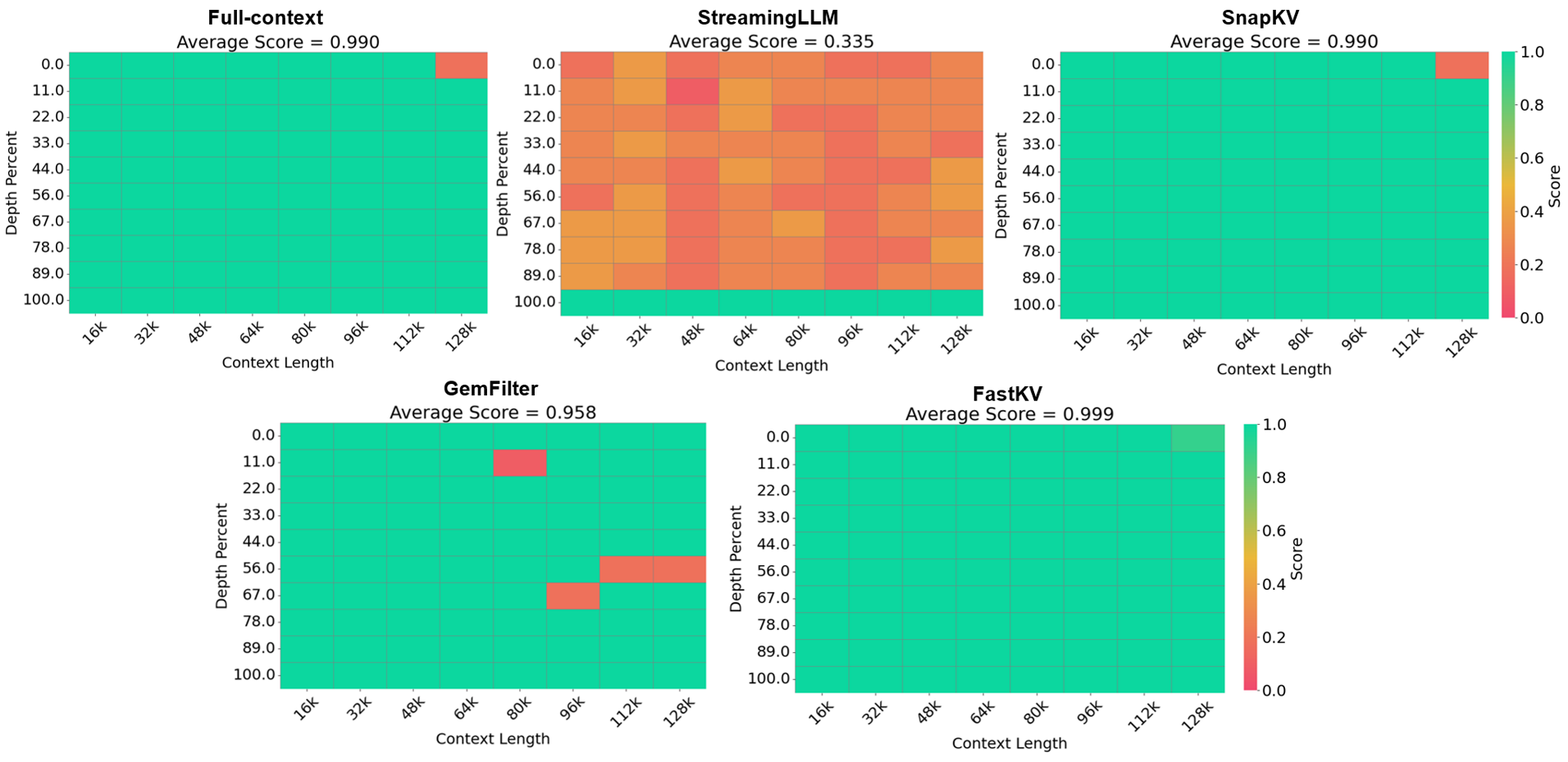}
    \caption{Needle-in-a-Haystack results of LLaMA-3.1-8B-Instruct with 10\% KV retention rate.}
    \label{fig8}
\end{figure*}

We provide full breakdown of Needle-in-a-Haystack results for LLaMA-3.1-8B-Instruct in Figure~\ref{fig8}.
The KV retention rate is fixed to 10\% for all methods, while GemFilter and FastKV reduces prefill compute rate to 51\% and 60\%, respectively.

FastKV achieves the highest score among all evaluated baselines.
FastKV even outperforms the full-context, as Token-Selective Propagation helps the model to focus on globally critical tokens by simplifying the input context.

\newpage
\subsection{Latency}

\begin{figure*}[h]
    \centering
    \includegraphics[width=0.95\textwidth]{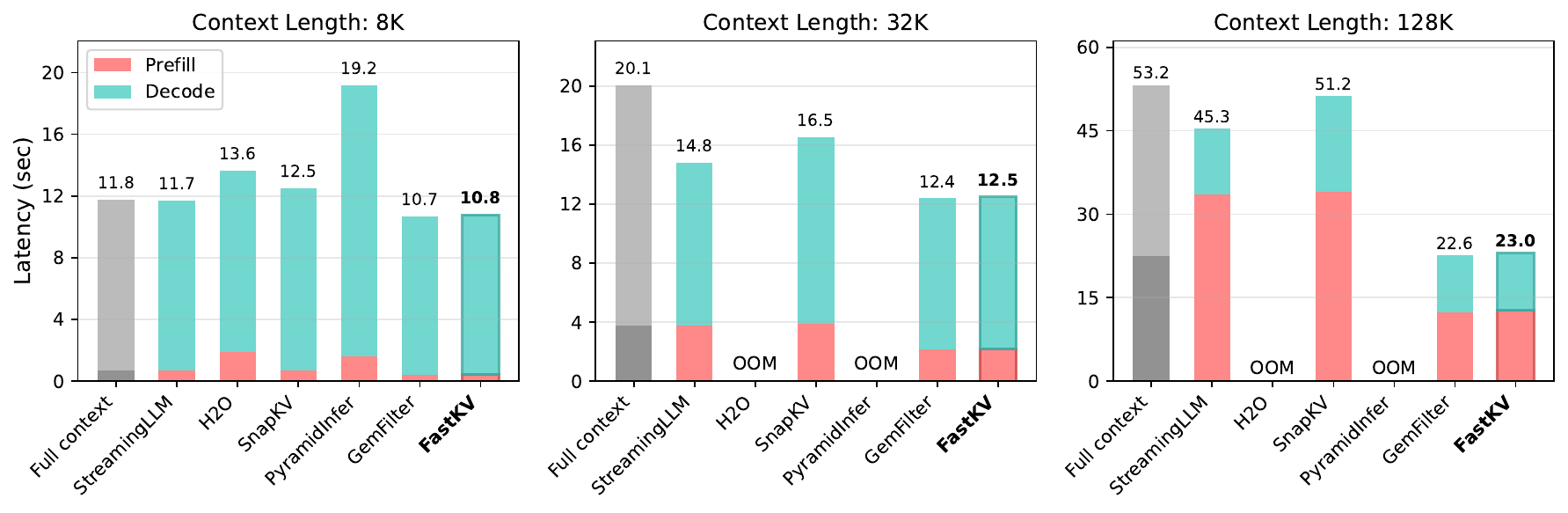}
    \caption{End-to-end inference latency breakdown of Ministral-8B-Instruct at varying input context lengths (generating 256 tokens).}
    \label{fig9}
\end{figure*}

We present the end-to-end inference latency breakdown of Ministral-8B-Instruct in Figure~\ref{fig9}.
The results were profiled with varying context lengths of 8K, 32K, and 128K tokens, while generation length is fixed to 256 tokens.

Prefill compute rate is set to 60\% for all prefill-aware methods, including PyramidInfer, GemFilter, and FastKV.
The KV retention rate is set to 10\% for all methods except PyramidInfer.
The KV retention rate of PyramidInfer is identical to its prefill compute rate, which is 60\%.

All of the experiments were conducted on a NVIDIA A100 SXM GPU.

The overall trend is consistent with the results observed on LLaMA-3.1-8B-Instruct.
As the context length increases, the end-to-end latency grows, with prefill latency accounting for an increasingly larger proportion of the total.

StreamingLLM and SnapKV effectively reduce decoding latency, but since they do not mitigate prefill latency, their benefits diminish in long-context scenarios where prefill dominates. 
H2O and PyramidInfer cannot utilize FlashAttention-2 due to implementation constraints, leading to out-of-memory errors beyond 8K tokens, and they are even slower than the full-KV baseline at 8K.
PyramidInfer exhibits particularly large decoding latency, as its KV retention rate is 60\%, forcing it to maintain a substantially larger KV cache than other methods that operate with only a 10\% budget.

By contrast, both GemFilter and FastKV address latency in both the prefill and decoding phases, achieving consistent acceleration across all settings and delivering over 2$\times$ speedup at a 128K context length. 
For LLaMA-3.1-8B-Instruct, FastKV was slightly slower than GemFilter because the filter layer (layer 13) precedes the TSP layer (layer 15). 
In Ministral, however, both the filter layer and the TSP layer are located at layer 17, resulting in identical latency for the two methods.

\newpage

\subsection{Token Importance Estimation Overhead Analysis}
\label{app:importance_overhead}

FastKV uses the same token importance estimation procedure for both Token-Selective Propagation and KV eviction.
Specifically, token importance is computed from a small recent attention window of 8 tokens, followed by MaxPooling
with kernel size 7 and uniform averaging across heads. Because the scoring is performed only over this short window,
its computational overhead is negligible compared to the overall prefill computation.

Table~\ref{tab:importance_overhead} reports the runtime breakdown of the token importance estimation step on
LLaMA-3.1-8B-Instruct. Even at 128K context length, the estimation step takes only 0.15 seconds, accounting for
0.88\% of the total prefill latency. This confirms that the token importance estimation used in FastKV introduces
only a small overhead while enabling both TSP and KV eviction.

\begin{table}[h]
  \centering
  \caption{Breakdown of token importance estimation overhead during prefill on LLaMA-3.1-8B-Instruct.}
  \label{tab:importance_overhead}
  \setlength{\tabcolsep}{6pt}
  \renewcommand{\arraystretch}{0.9}
  \scalebox{0.88}{
  \begin{tabular}{lccc}
    \toprule
    Context Length & Prefill (sec) & Estimation (sec) & Overhead / Latency \\
    \midrule
    32K  & 2.10 $\pm$ 0.002  & 0.04 $\pm$ 0.0001 & 1.90\% \\
    64K  & 5.55 $\pm$ 0.013  & 0.08 $\pm$ 0.0002 & 1.44\% \\
    128K & 16.99 $\pm$ 0.019 & 0.15 $\pm$ 0.0002 & 0.88\% \\
    \bottomrule
  \end{tabular}
  }
\end{table}

\subsection{Hyperparameter Analysis}
\label{app:hyperparameter_analysis}

We provide additional hyperparameter sweeps of FastKV on LLaMA-3.1-8B-Instruct to show the effect of various design choices.

\paragraph{TSP Rate and KV Retention Rate.}
Table~\ref{tab:tsp_kv_sweep} presents a 2D sweep over TSP rate and KV retention rate on
LLaMA-3.1-8B-Instruct. The results show that the two hyperparameters affect accuracy largely
independently, with only limited interaction across the grid. This supports our design choice of
decoupling prefill reduction through TSP from decoding-time KV retention. The configurations used
in the main paper (TSP rate = 0.2, KV retention rate = 0.1/0.2) achieve a reasonable
accuracy--latency trade-off.

\begin{table}[h]
  \centering
  \caption{2D sweep over TSP rate and KV retention rate on LLaMA-3.1-8B-Instruct.}
  \label{tab:tsp_kv_sweep}
  \setlength{\tabcolsep}{6pt}
  \renewcommand{\arraystretch}{0.9}
  \scalebox{0.88}{
  \begin{tabular}{lccccc}
    \toprule
    \textbf{TSP Rate $\backslash$ KV Retention Rate} & \textbf{0.1} & \textbf{0.2} & \textbf{0.3} & \textbf{0.4} & \textbf{0.5} \\
    \midrule
    0.1 & 47.80 & -     & -     & -     & -     \\
    0.2 & 48.47 & 49.07 & -     & -     & -     \\
    0.3 & 48.55 & 49.25 & 49.46 & -     & -     \\
    0.4 & 48.52 & 49.30 & 49.48 & 49.65 & -     \\
    0.5 & 48.54 & 49.38 & 49.60 & 49.67 & 49.86 \\
    \bottomrule
  \end{tabular}
  }
\end{table}

\paragraph{TSP Rate and TSP Layer Index.}
To more systematically evaluate the trade-off frontier of TSP and quantify how TSP choices affect
accuracy, we additionally sweep over a wide range of TSP layer indices and TSP rates on
LLaMA-3.1-8B-Instruct. Table~\ref{tab:tsp_layer_sweep} provides the full trade-off surface.
The results show that applying TSP too early or using an excessively low TSP rate significantly
degrades accuracy, while moving to later TSP layers or using larger TSP rates yields only marginal
additional gains. The configuration used in the main paper (TSP rate = 0.2, TSP layer = 15) lies in
a broad Pareto-optimal region, offering near-minimal latency without compromising accuracy.

\begin{table}[h]
  \centering
  \caption{2D sweep over TSP rate and TSP layer index on LLaMA-3.1-8B-Instruct.}
  \label{tab:tsp_layer_sweep}
  \setlength{\tabcolsep}{6pt}
  \renewcommand{\arraystretch}{0.9}
  \scalebox{0.88}{
  \begin{tabular}{lccccccc}
    \toprule
    \textbf{TSP Rate $\backslash$ TSP Layer} & \textbf{0} & \textbf{5} & \textbf{10} & \textbf{15} & \textbf{20} & \textbf{25} & \textbf{30} \\
    \midrule
    0.1 & 30.43 & 38.29 & 46.62 & 47.80 & 48.18 & 48.26 & 48.34 \\
    0.2 & 34.18 & 41.85 & 48.11 & 48.47 & 48.41 & 48.26 & 48.32 \\
    0.3 & 35.57 & 44.24 & 48.35 & 48.55 & 48.36 & 48.35 & 48.34 \\
    0.4 & 37.40 & 45.44 & 48.36 & 48.52 & 48.40 & 48.45 & 48.42 \\
    0.5 & 40.62 & 46.31 & 48.53 & 48.54 & 48.46 & 48.44 & 48.39 \\
    \bottomrule
  \end{tabular}
  }
\end{table}

\end{document}